\documentclass[letterpaper, 10 pt, conference]{ieeeconf}  % Comment this line out if you need a4paper

\IEEEoverridecommandlockouts                              % This command is only needed if 
\usepackage{graphics} % for pdf, bitmapped graphics files
\usepackage{epsfig} % for postscript graphics files
\usepackage{mathptmx} % assumes new font selection scheme installed
\usepackage{times} % assumes new font selection scheme installed
\usepackage{amsmath} % assumes amsmath package installed
\usepackage{amssymb}  % assumes amsmath package installed\usepackage{graphicx}
\usepackage{graphicx}
\usepackage{booktabs} % for professional tables
\usepackage{algorithm}
\usepackage{algorithmic}
\usepackage{color,soul}
\usepackage{url,hyperref}

\title{\LARGE \bf
RetinaGAN: An Object-aware Approach to Sim-to-Real Transfer
}

\author{
       Daniel Ho$^{1}$,
       Kanishka Rao$^{2}$,
       Zhuo Xu$^{3}$,
       Eric Jang$^{2}$,
       Mohi Khansari$^{1}$,
       Yunfei Bai$^{1}$% <-this % stops a space

\thanks{$^{1}$Everyday Robots, X The Moonshot Factory, Mountain View, CA, USA}%
\thanks{$^{2}$Robotics at Google, Mountain View, CA, USA}%
\thanks{$^{3}$University of California, Berkeley, Berkeley, CA, USA}%
}

\usepackage{etoolbox}
\makeatletter
\patchcmd{\@makecaption}
  {\scshape}
  {}
  {}
  {}
\makeatother

\begin{document}

\maketitle
\thispagestyle{empty}
\pagestyle{empty}

%%%%%%%%%%%%%%%%%%%%%%%%%%%%%%%%%%%%%%%%%%%%%%%%%%%%%%%%%%%%%%%%%%%%%%%%%%%%%%%%
\begin{abstract}
The success of deep reinforcement learning (RL) and imitation learning (IL) in vision-based robotic manipulation typically hinges on the expense of large scale data collection. With simulation, data to train a policy can be collected efficiently at scale, but the visual gap between sim and real makes deployment in the real world difficult. We introduce RetinaGAN, a generative adversarial network (GAN) approach to adapt simulated images to realistic ones with object-detection consistency. RetinaGAN is trained in an unsupervised manner without task loss dependencies, and preserves general object structure and texture in adapted images. We evaluate our method on three real world tasks: grasping, pushing, and door opening. RetinaGAN improves upon the performance of prior sim-to-real methods for RL-based object instance grasping and continues to be effective even in the limited data regime. When applied to a pushing task in a similar visual domain, RetinaGAN demonstrates transfer with no additional real data requirements. We also show our method bridges the visual gap for a novel door opening task using imitation learning in a new visual domain. Visit the project website at \url{retinagan.github.io}

\end{abstract}

%%%%%%%%%%%%%%%%%%%%%%%%%%%%%%%%%%%%%%%%%%%%%%%%%%%%%%%%%%%%%%%%%%%%%%%%%%%%%%%%

\section{Introduction}

\begin{figure}[thpb]
  \centering
  \includegraphics[width=0.97\linewidth]{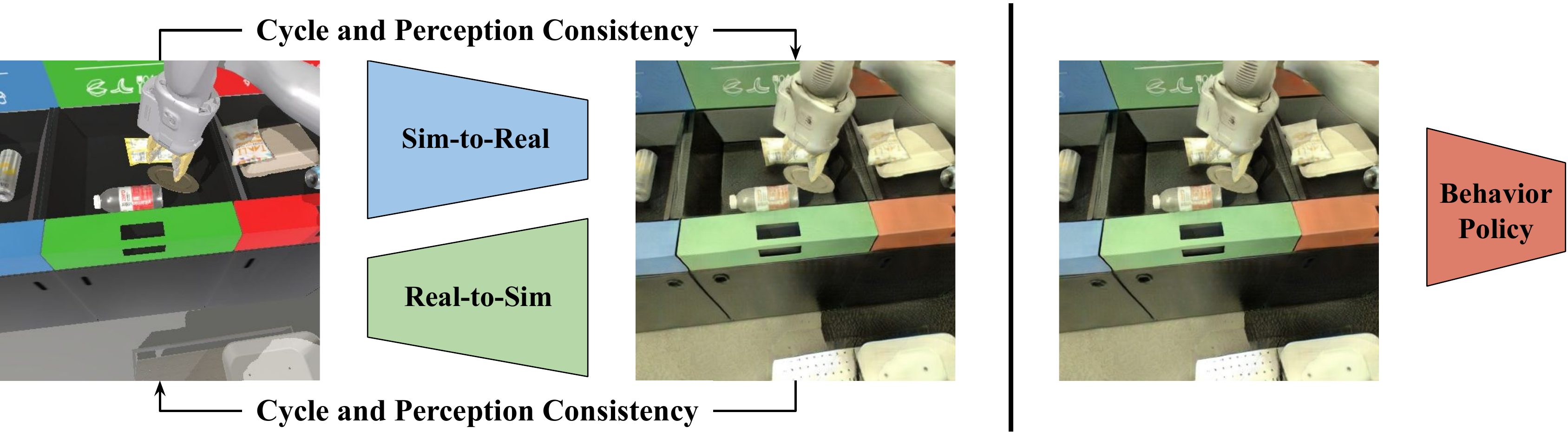}
  \caption{Overview of RetinaGAN pipeline. Left: Train RetinaGAN using pre-trained perception model to create a sim-to-real model. Right: Train the behavior policy model using the sim-to-real generated images. This policy can later be deployed in real.}
  \label{fig:gan1}
\end{figure}

Vision-based reinforcement learning and imitation learning methods incorporating deep neural network structure can express complex behaviors, and they solve robotics manipulation tasks in an end-to-end fashion \cite{levine2018learning,pinto2016supersizing,kalashnikov2018qtopt}. These methods are able to generalize and scale on complicated robot manipulation tasks, though they require many hundreds of thousands of real world episodes which are costly to collect.

Some of this data collection effort can be mitigated by collecting these required episodes in simulation and applying sim-to-real transfer methods. Simulation provides a safe, controlled platform for policy training and development with known ground truth labels. Such simulated data can be cheaply scaled. However, directly executing such a policy in the real world typically performs poorly, even if the simulation configuration is carefully controlled, because of visual and physical differences between the domains known as the reality gap. In practice, we find the visual difference to be the bottleneck in our learning algorithms and focus further discussion solely on this.

One strategy to overcome the visual reality gap is pixel-level domain adaptation; such methods may employ generative adversarial networks to translate the synthetic images to the real world domain \cite{bousmalis2017unsupervised}. However, a GAN may arbitrarily change the image, including removing information necessary for a given task. More broadly for robotic manipulation, it is important to preserve scene features that directly interact with the robot, like object-level structure and textures.

To address this, we propose RetinaGAN, a domain adaptation technique which requires strong object semantic awareness through an object detection consistency loss. RetinaGAN involves a CycleGAN \cite{zhu2020unpaired} that adapts simulated images to look more realistic while also resulting in consistent objects predictions.
We leverage an object detector trained on both simulated and real domains to make predictions on original and translated images, and we enforce the invariant of the predictions with respect to the GAN translation.

RetinaGAN is a general approach to adaptation which provides reliable sim-to-real transfer for tasks in diverse visual environments (Fig. \ref{fig:gan1}).
In a specific scenario, we show how RetinaGAN may be reused for a novel pushing task.
We evaluate the performance of our method on three real world robotics tasks and demonstrate the following:

\begin{enumerate}

\item  RetinaGAN, when trained on robotic grasping data, allows for grasping RL task models that outperform prior sim-to-real methods on real world grasping by 12\%.

\item  With limited (5-10\%) data, our method continues to work effectively for grasping, only suffering a 14\% drop in performance.

\item The RetinaGAN trained with grasping data may be reused for another similar task, 3D object pushing, without any additional real data. It achieves 90\% success.

\item We train RetinaGAN for a door opening imitation learning task in a drastically different environment, and we introduce an Ensemble-RetinaGAN method that adds more visual diversity to achieve 97\% success rate.

\item We utilize the same pre-trained object detector in all experiments.

\end{enumerate}

\section{Related Work}

To address the visual sim-to-reality gap, prior work commonly apply domain randomization and domain adaptation techniques.

With domain randomization, a policy is trained with randomized simulation parameters and scene configurations which produce differences in visual appearance \cite{8202133,matas2018simtoreal,James2017TransferringEV,8578591,yan2019data,46985}. The policy may learn to generalize across the parameter distribution and takes actions likely to work in all situations. Policy performance relies heavily on the kind of randomizations applied and whether they are close to or cover reality. The recently proposed method, Automatic Domain Randomization \cite{akkaya2019solving}, automates the hyperparameter tuning process for Rubik's Cube manipulation. But, domain randomization still requires manual, task-specific selection of visual parameters like the scene, textures, and rendering.

Domain adaptation bridges the reality gap by directly resolving differences between the domains \cite{7078994}. Images from a source domain can be modified at the pixel-level to resemble a target domain \cite{bousmalis2017unsupervised,yoo2016pixellevel}. Or, feature-level adaptation aligns intermediate network features between the domains \cite{6126344,pmlr-v37-long15,fang2018multi}. GANs are a commonly applied method for pixel-level transfer which only require unpaired images from both domains \cite{NIPS2014_5423,brock2019large,zhu2020unpaired,bousmalis2018using,james2019sim}. Our method employs such pixel-level adaptation to address the sim-to-real gap.

Action Image \cite{khansari2020action} is another approach to bridge the sim-to-real gap through learning a domain invariant representation for the task of grasping. Our work is complementary to this work and can help to further reduce this gap.

Among prior work that apply semantic consistency to GAN training, CyCADA \cite{hoffman2017cycada} introduces a pixel-level perception consistency loss (semantic segmentation) as a direct task loss, and applies the learned generator to other semantic segmentation and perception tasks. Comparatively, RetinaGAN uses object detection where labels on real data is much easier to obtain and demonstrates that feature understanding from object detection is sufficient to preserve object semantics for robotics applications.

Recently, RL-CycleGAN \cite{rao2020rlcyclegan} extends vanilla CycleGAN \cite{zhu2020unpaired} with an additional reinforcement learning task loss. RL-CycleGAN enforces consistency of task policy Q-values between the original and transferred images to preserve information important to a given task. RL-CycleGAN is trained jointly with the RL model and requires task-specific real world episodes collected via some preexisting policy. Comparatively, RetinaGAN works for supervised and imitation learning, as it uses object detection as a task-decoupled surrogate for object-level visual domain differences. This requires additional real-world bounding box labels, but the detector can be reused across robotics tasks. In practice, we find the RetinaGAN easier to train since the additional object detector is pre-trained and not jointly optimized.

\section{Preliminaries}

\subsection{Object Detection}
\begin{figure}[thpb]
  \centering
  \includegraphics[width=1.0\linewidth]{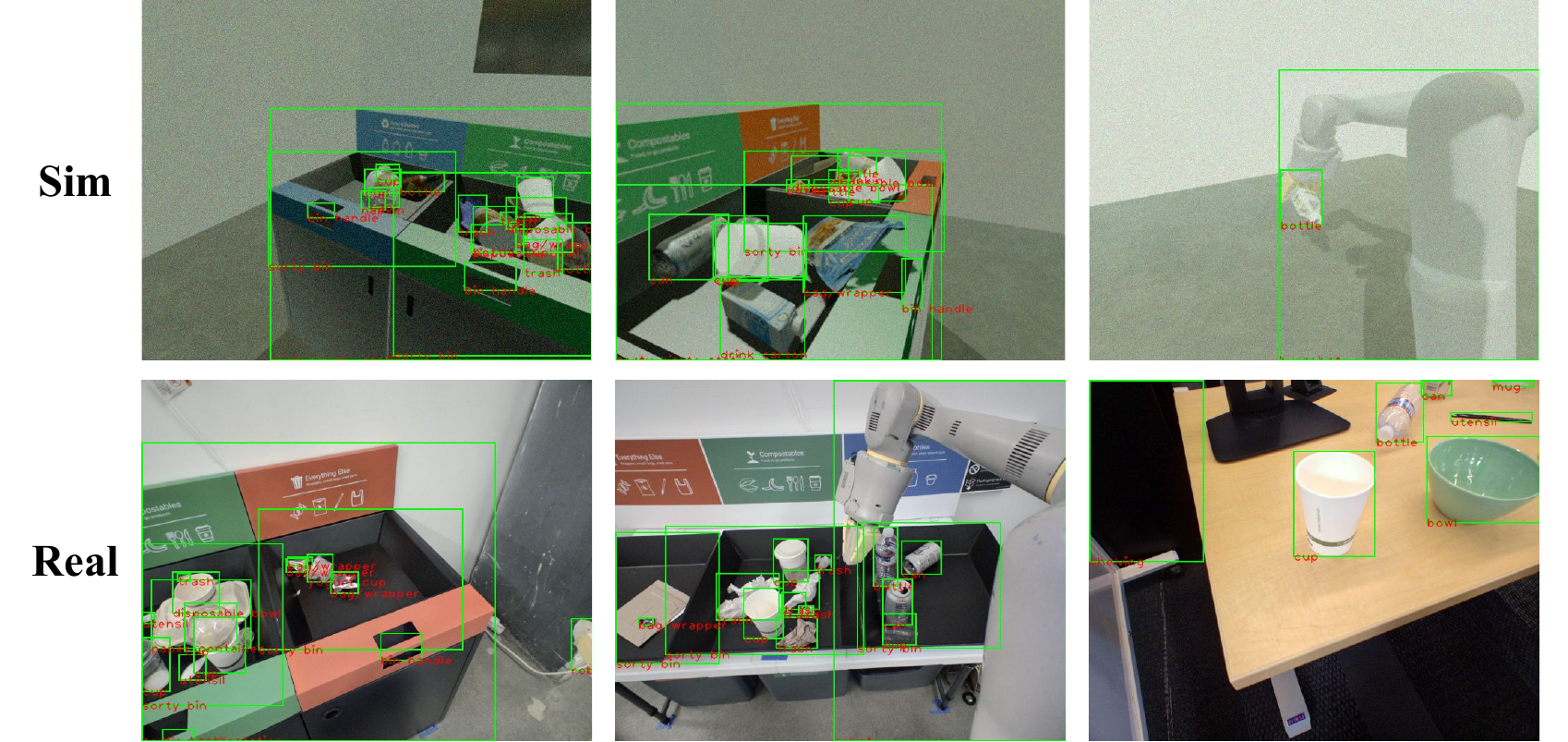}
  \caption{Sim and real perception data used to train EfficientDet focuses on scenes of disposable objects encountered in recycling stations. The real dataset includes 44,000 such labeled images and 37,000 images of objects on desks. The simulated dataset includes 625,000 total images.}
  \label{fig:prcp}
\end{figure}

We leverage an object detection perception model to provide object awareness for the sim-to-real CycleGAN. We train the model by mixing simulated and real world datasets which contain ground-truth bounding box labels (illustrated in Fig. \ref{fig:prcp}). The real world object detection dataset includes robot images collected in general robot operation; labeling granularity is based on general object type -- all brands of soda will be part of the ``can'' class.
Simulation data is generated with the PyBullet physics engine \cite{coumans2017pybullet}.

Object detection models are object-aware but task-agnostic, and thus, they do not require task-specific data. We use this single detection network as a multi-domain model for all tasks, and we suspect in-domain detection training data is not crucial to the success of our method. Notably, the door opening domain is very different from the perception training data domain, and we demonstrate successful transfer in Section \ref{sec:door_opening}.

While the initial dataset required for object detection can be a significant expense, leveraging off-the-shelf models is a promising direction, especially given our experimental results with door opening. 
% pre-training from benchmark datasets and transfer learning are promising, well-studied techniques that reduce such cost.
Furthermore, detection is a generally useful robot capability, so roboticists may create detection datasets for use cases beyond sim-to-real.

We select the EfficientDet-D1 \cite{efficientdet} model architecture (using the same losses as  RetinaNet \cite{retinanet}) for the object detector. EfficientDet passes an input RGB image through a backbone feedforward EfficientNet \cite{tan2020efficientnet} architecture, and fuses features at multiple scales within a feature pyramid network. From the result, network heads predict class logit and bounding box regression targets.

% We note that it is also possible to train separate perception networks for each domain. However, this adds complexity and requires that the object sets between synthetic and real data be close to bijective, because both models would have to produce consistent predictions on perfectly paired images.

% While segmentation models like Mask-RCNN \cite{he2018mask} and ShapeMask \cite{kuo2019shapemask} provide dense, pixel-level object supervision, it is practically easier and more efficient to label object detection data. In general, perception models are common, and data collection and labeling efforts can be amortized across various use cases.

\subsection{CycleGAN}

The RetinaGAN training process builds on top of CycleGAN \cite{zhu2020unpaired}: an approach to learn a bidirectional mapping between unpaired datasets of images from two domains, $X$ and $Y$, with generators $G: X \rightarrow Y$ and $F: Y \rightarrow X$. These generators are trained alongside adversarial discriminators $D_x, D_y$, which classify images to the correct domain, and with the cycle consistency loss capturing $F(G(x)) \approx x, G(F(y) \approx y$ for $x \in X, y \in Y$. We can summarize the training process with the CycleGAN loss (described in detail in \cite{zhu2020unpaired,rao2020rlcyclegan}):
\begin{align}
    \mathcal{L}_\text{CycleGAN}(G, F, D_x, D_y) &=  \mathcal{L}_\text{GAN}(G, D_Y, X, Y) \nonumber \\
    &+ \mathcal{L}_\text{GAN}(F, D_X, Y, X)  \\
    &+ \lambda_\text{cycle}\mathcal{L}_\text{cycle}(G, F) \nonumber
\end{align}

\section{RetinaGAN}

\begin{algorithm}[tb]
  \caption{Summary of RetinaGAN training pipeline.}
  \label{alg:apply}
\begin{algorithmic}[1]
\vspace{.1cm}
%   \STATE {\bfseries Input:} \textbf{RetinaNet} $R$, \textbf{dataset} $D$ %, [list of $(op, prob, mag)$]
  \STATE \textbf{Given:} EfficientDet, $Det$, trained with simulation and real robot data
  \STATE Collect simulation ($X$) and real ($Y$) task episodes
  \WHILE{train $G: X \rightarrow Y$ and $F: Y \rightarrow X$ generators}
  \STATE Iterate over batch of simulation ($x$) and real ($y$) data
  \STATE Compute $G(x) = x'$, $F(x') = x'', F(y) = y', G(y') = y''$
  \FOR{pairs $p_1, p_2$ in \{x, x', x''\}, \{y, y', y''\}}
      \STATE Compute $Det(p_1) \approx Det(p_2)$ loss, $\mathcal{L}_\text{prcp}(p_1, p_2)$
  \ENDFOR
  \STATE Compute CycleGAN losses, $\mathcal{L}_\text{CycleGAN}$
  \STATE Take optimization step using losses
  \ENDWHILE
  %\STATE Train task policy, adapting simulation images using $G$
  %\STATE Run task policy inference on real images
\end{algorithmic}
\end{algorithm}

\begin{figure}[thpb]
  \centering
  \includegraphics[width=.9\linewidth]{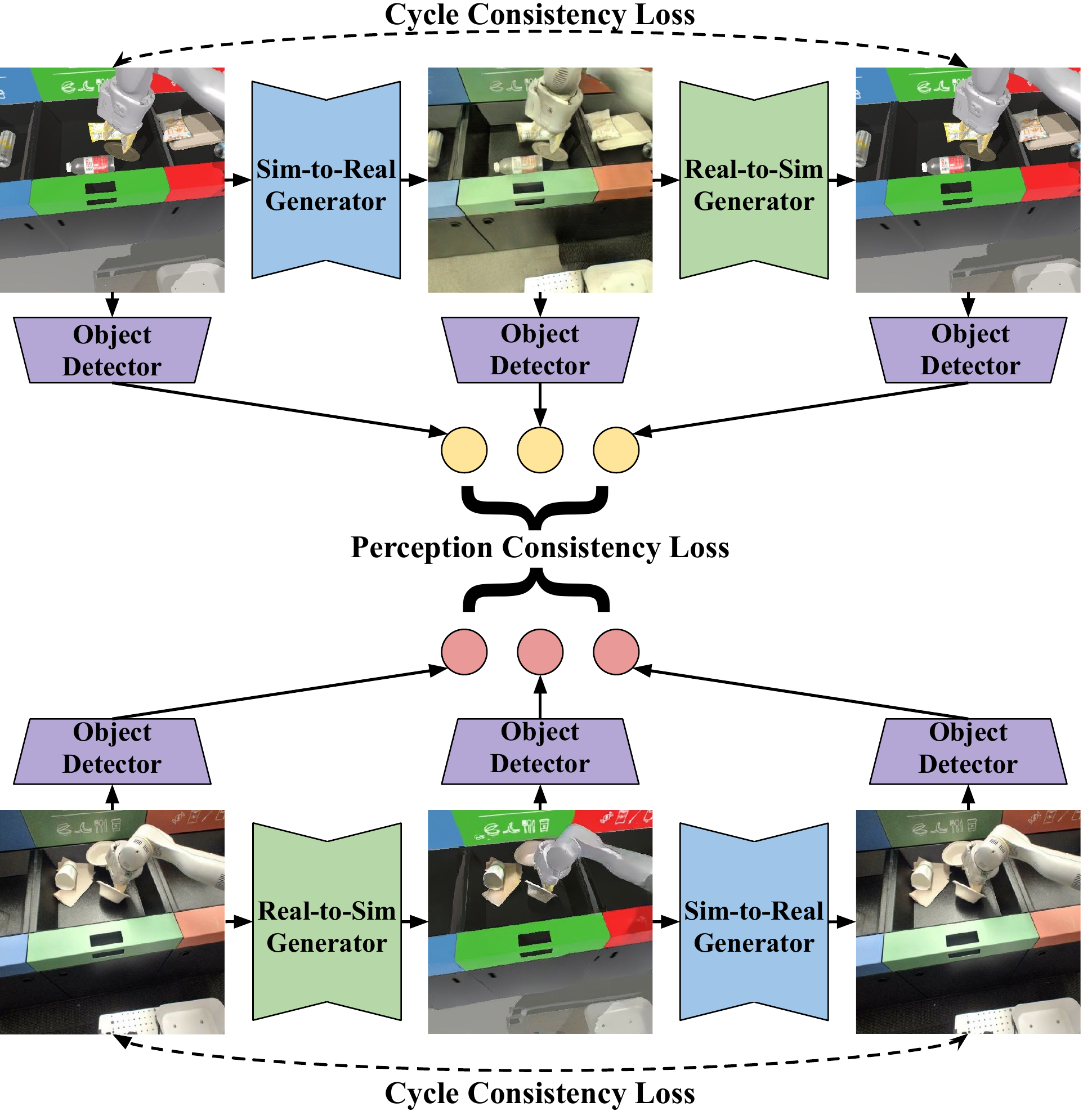}
  \caption{Diagram of RetinaGAN stages.
  The simulated image (top left) is transformed by the sim-to-real generator and subsequently by the real-to-sim generator. The perception loss enforces consistency on object detections from each image. The same pipeline occurs for the real image branch at the bottom.}
  \label{fig:gan3}
\end{figure}

RetinaGAN trains with a frozen object detector, EfficientDet, that provides object consistency loss. Once trained, the RetinaGAN model adapts simulated images for training the task policy model. Similarly to CycleGAN, we use unpaired data without labels. The overall framework is described in Algorithm \ref{alg:apply} and illustrated in Fig. \ref{fig:gan3}, and the details are described below.

\begin{figure}[thpb]
  \centering
  \includegraphics[width=1.0\linewidth]{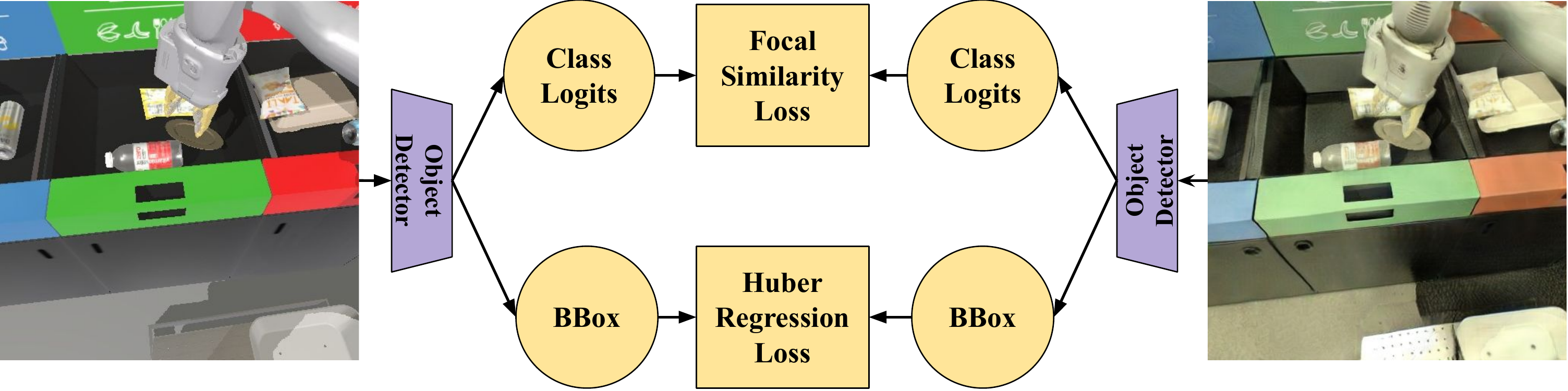}
  \caption{Diagram of perception consistency loss computation. An EfficientDet object detector predicts boxes and classes. Consistency of predictions between images is captured by losses similar to those in object detection training.}
  \label{fig:gan2}
\end{figure}

From CycleGAN, we have six images: sim, transferred sim, cycled sim, real, transferred real, and cycled real. Because of object invariance with respect to transfer, an oracle domain adapter would produce identical predictions between the former three images, as well as the latter three. To capture this invariance, we run inference using a pre-trained and frozen EfficientDet model on each image; for each of these pairs, we compute a perception consistency loss.

\subsection{Perception Consistency Loss}
The perception consistency loss penalizes the generator for discrepancies in object detections between translations. Given an image $I$, EfficientDet predicts a series of anchor-based bounding box regressions and class logits at several levels in its Feature Pyramid Network \cite{lin2017feature}.

We compute the perception consistency loss ($\mathcal{L}_\text{prcp}$) given a pair of images similarly to the box and class losses in typical RetinaNet/EfficientDet training.
However, because the Focal Loss \cite{retinanet}, used as the class loss, assumes one-hot vector ground truth labels, we propose a variation called Focal Consistency Loss (FCL) which is compatible with logit/probability labels (explained below in Section \ref{sec:fce}).

Without loss of generality, consider an image pair to be $x$ and $G(x)$. This loss can be computed with a pre-trained EfficientDet network as:
\begin{align}
\text{box}_{x}, \text{cls}_{x} &= \text{EfficientDet}(x) \\
\text{box}_{G(x)}, \text{cls}_{G(x)} &= \text{EfficientDet}(G(x)) \\
\mathcal{L}_\text{prcp}(x, G(x)) &= \mathcal{L}_\text{Huber}(\text{box}_{x}, \text{box}_{G(x)}) \\ &+ \text{FCL}(\text{cls}_{x}, \text{cls}_{G(x)}) \nonumber
\end{align}
$\mathcal{L}_\text{Huber}$ is the Huber Loss \cite{elsr} used as the box regression loss. This process is visualized in Fig. \ref{fig:gan2}.

The Perception Consistency Loss on a batch of simulated images $x$ and real images $y$, using the sim-to-real generator $G$ and the real-to-sim generator $F$, is:
\begin{align}
\mathcal{L}_\text{prcp}(x,y,F,G) &= \mathcal{L}_\text{prcp}(x, G(x)) + \frac{1}{2} \mathcal{L}_\text{prcp}(x, F(G(x))) \nonumber \\
            &+ \frac{1}{2} \mathcal{L}_\text{prcp}(G(x), F(G(x)) \\
            &+ \mathcal{L}_\text{prcp}(y, F(y)) \nonumber + \frac{1}{2} \mathcal{L}_\text{prcp}(y, G(F(y))) \nonumber \\
            &+ \frac{1}{2} \mathcal{L}_\text{prcp}(F(y), G(F(y)) \nonumber
\end{align}
We halve the losses involving the cycled $F(G(x))$ and $G(F(x))$ images because they are compared twice (against the orginal and transferred images), but find that this weight has little effect in practice.

We arrive at the overall RetinaGAN loss:
\begin{align}
    \mathcal{L}_\text{RetinaGAN}(G, F, D_x, D_y) &= \mathcal{L}_\text{CycleGAN}(G, F, D_x, D_y) \\
    &+ \lambda_\text{prcp}\mathcal{L}_\text{prcp}(x,y,F,G) \nonumber
\end{align}
\subsection{Focal Consistency Loss (FCL)}
\label{sec:fce}
We introduce and derive a novel, interpolated version of the Focal Loss (FL) called Focal Consistency Loss (FCL), which extends support to a ground truth confidence probability $y \in [0,1]$ from a binary $y \in \{0, 1\}$. Focal losses handle class imbalances in one-stage object detectors, improving upon Cross Entropy (CE) and Balanced Cross Entropy (BCE) losses (Section 3, \cite{retinanet}).

We begin from CE loss, which can be defined as:
\begin{equation}
\text{CE}(y,p) = y \log p - (1-y)\log(1-p)
\end{equation}
where $p$ is the predicted probability.

BCE loss handles class imbalance by including a weighting term $\alpha \in [0,1]$ if $y=1$ and $1 - \alpha$ if $y = 0$. Interpolation between these two terms yields:
\begin{equation}
\text{BCE}(y,p) = [(2\alpha-1)p+(1-\alpha)]\text{CE}(y,p)
\end{equation}
Focal Loss weights BCE by a focusing factor of $(1-p_t)^\gamma$, where $\gamma \ge 0$ and $p_t$ is $p$ if $y = 0$ and $1-p$ if $y = 1$ to addresses foreground-background imbalance. FCL is derived through interpolation between the binary cases of $p_t$:
\begin{equation}
\text{FCL}(y,p) = \left|y-p\right|^\gamma \text{BCE}(y,p)
\end{equation}
FCL is equivalent to FL when the class targets are one-hot labels, but interpolates the loss for probability targets.
Finally, FL is normalized by the number of anchors assigned to ground-truth boxes (Section 4, \cite{retinanet}). Instead, FCL is normalized by the total probability attributed to anchors in the class tensor. This weights each anchor by its inferred probability of being a ground-truth box.

\subsection{Hyperparameters}
We follow the hyperparameter selection of $\lambda_\text{cycle}=10$ from RL-CycleGAN without tuning. $\lambda_\text{prcp}$ trades focus on object reconstruction quality with overall image realism. We find $0.1$ to $1.0$ to be stable, and selected $0.1$ for all experiments, as objects were well-preserved at this value. We find relative weights between $\mathcal{L}_\text{prcp}$ terms not important.

\section{Task Policy Models and Experiments}
We aim to understand the following scenarios: 1) the value of sim-to-real at various data sizes by comparing robotics models trained with RetinaGAN vs without RetinaGAN 2) with purely sim-to-real data, how models trained with various GANs perform 3) transfer to other tasks.

We begin with training and evaluating RetinaGAN for RL grasping. We then proceed by applying the same RetinaGAN model to RL pushing and finally re-train on an IL door opening task. See the Appendix for further details on training and model architecture.

\subsection{Reinforcement Learning: Grasping}

\begin{figure*}[thpb]
  \centering
  \includegraphics[width=.95\linewidth]{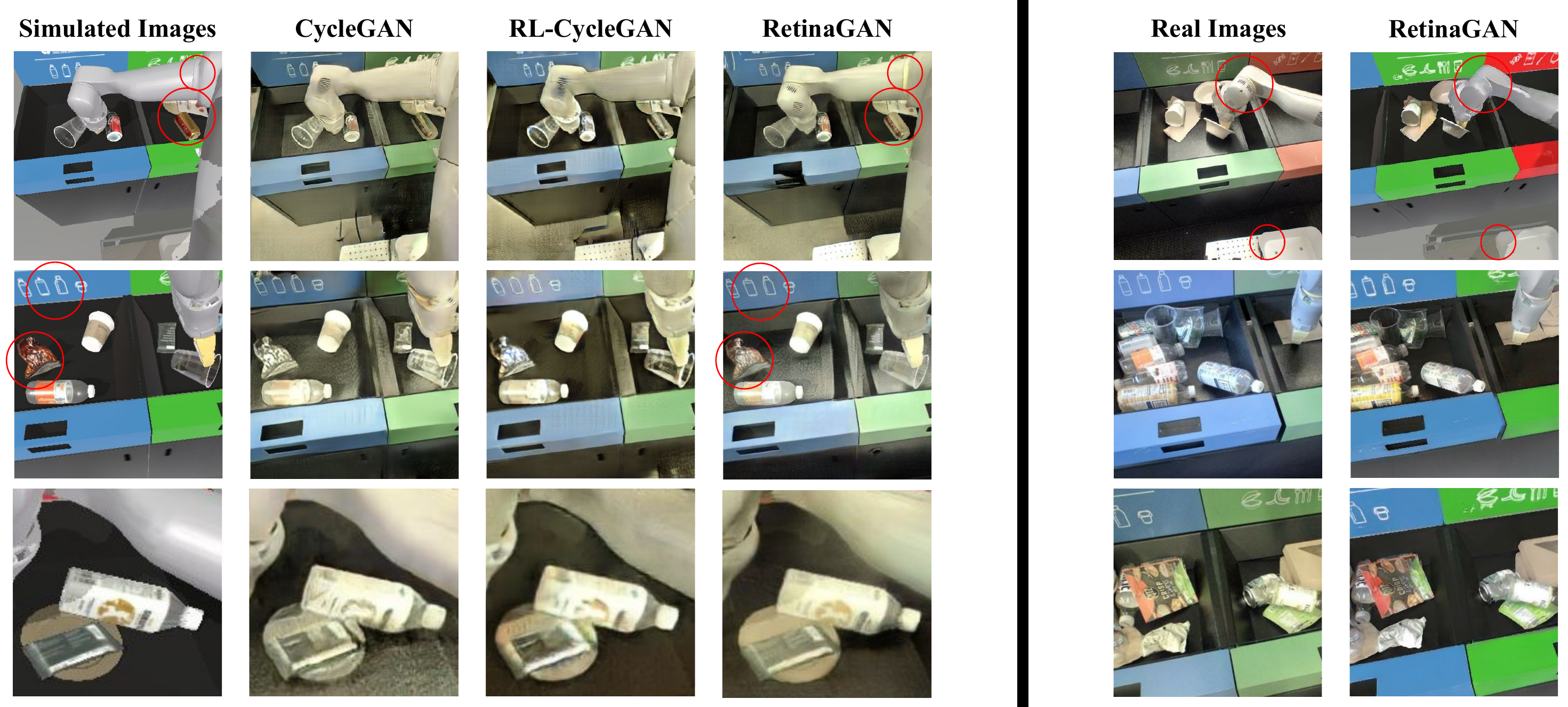}
  \caption{
  Sampled, unpaired images for the grasping task at various scales translated with either the sim-to-real (left) or real-to-sim (right) generator. Compared to other methods, the sim-to-real RetinaGAN consistently preserves object textures and better reconstructs real features. The real-to-sim RetinaGAN is able to preserve all object structure in cluttered scenes, and it correctly translates details of the robot gripper and floor.
  }
  \label{fig:grasping}
\end{figure*}

We use the distributed reinforcement learning method Q2-Opt \cite{bodnar2019quantile}, an extension to QT-Opt \cite{kalashnikov2018qtopt}, to train a vision based task model for instance grasping. In the grasping task, a robot is positioned in front of one of three bins within a trash sorting station and attempts to grasp targeted object instances. The RGB image and a binary mask for the grasp target is input into the network. Real world object classes are focused on cups, cans, and bottles, although real training data is exposed to a long tail of discarded objects. Grasps in simulation are performed with the PyBullet \cite{coumans2017pybullet} physics engine, with 9 to 18 spawned objects per scene. Example images are visualized in Fig. \ref{fig:grasping}.

When using real data, we train RetinaGAN on 135,000 off-policy real grasping episodes and the Q2-Opt task model on 211,000 real episodes. We also run a low data experiment using 10,000 real episodes for training both RetinaGAN and Q2-Opt. We run distributed simulation to generate one-half to one million on-policy training episodes for RetinaGAN and one to two million for Q2-Opt.

We evaluate with six robots and sorting stations. Two robots are positioned in front of each of the three waste bins, and a human manually selects a cup, can, or bottle to grasp. Each evaluation includes thirty grasp attempts for each class, for ninety total. By assuming each success-failure experience is an independent Bernouili trial, we can estimate the sample standard deviation as $\sqrt{q(1-q)/(n-1)}$, where $q$ is the average failure rate and $n$ is the number of trials.

\begin{table}[thpb]

\caption{Instance grasping success mean and estimated standard deviation (est. std.) of Q2-Opt compared between different training data sources across 90 trials. Results are organized by the number of real grasping episodes used.
}

\label{tab:rl}
\begin{center}

\begin{tabular}{lcc}

\toprule
\textbf{Model} & \textbf{Grasp Success} & \textbf{Est. Std.} \\

\midrule
Sim-Only         & 18.9\% & 4.1\% \\
Randomized Sim   & 41.1\% & 5.2\% \\

\midrule
\textbf{GAN: 10K Real, Q2-Opt: 10K Real}    & \\

Real & 22.2\% & 4.4\% \\
RetinaGAN & 47.4\% & 5.3\% \\
RetinaGAN+Real & \textbf{65.6\%} & 5.0\% \\

\midrule
\textbf{GAN: 135K Real, Q2-Opt: 211K Real}    & \\

Real                  & 30.0\% & 4.9\% \\
Sim+Real              & 54.4\% & 5.3\% \\
RetinaGAN+Real        & \textbf{80.0\%} & 4.2\% \\

\midrule
\textbf{GAN: 135K Real, Q2-Opt: 0 Real}    & \\
CycleGAN \cite{zhu2020unpaired}  & 67.8\% & 5.0\% \\
RL-CycleGAN \cite{rao2020rlcyclegan}  & 68.9\% & 4.9\% \\
RetinaGAN        & \textbf{80.0\%} & 4.2\% \\

\bottomrule
\end{tabular}
\end{center}
\end{table}

We use the RL grasping task to measure the sim-to-real gap and compare methods in the following scenarios, which are displayed in Table \ref{tab:rl}:

\begin{itemize}
    \item Train by mixing 10K real episodes with simulation to gauge data efficiency in the limited data regime.
    \item Train by mixing 135K+ real grasping episodes with simulation to investigate scalability with data, data efficiency, and performance against real data baselines.
    \item Train Q2-Opt with only simulation to compare between RetinaGAN and other sim-to-real methods.
\end{itemize}

In the sim-only setup, we train with fixed light position and object textures, though we apply photometric distortions including brightness, saturation, hue, contrast, and noise. In simulation evaluation, a Q2-Opt model achieves 92\% instance grasping success on cups, cans, and bottles. A performance of 18.9\% on the real object equivalents indicates a significant sim-to-real gap from training in simulation alone.

We compare against baselines in domain randomization and domain adaptation techniques. Domain randomization includes variations in texture and light positioning.

On the limited 10K episode dataset, RetinaGAN+Real achieves 65.6\%, showing significant performance improvement compared to Real-only.
When training on the large real dataset, RetinaGAN achieves 80\%, demonstrating scalability with more data. Additionally, we find that RetinaGAN+Real with 10K examples outperforms Sim+Real with 135K+ episodes, showing more than 10X data efficiency.

We proceed to compare our method with other domain adaptation methods; here, we train Q2-Opt solely on sim-to-real translated data for a clear comparison. RL-CycleGAN is trained with the same indiscriminate grasping task loss as in \cite{rao2020rlcyclegan}, but used to adapt on instance grasping. This could explain its relatively lower improvement from results in \cite{rao2020rlcyclegan}. RetinaGAN achieves 80\%, outperforming other methods by over two standard deviations, and interestingly, is on par with RetinaGAN+Real. We hypothesize that the knowledge of the real data was largely captured during RetinaGAN training, and the near on-policy simulation data is enough to train a high performing model.

\subsection{Reinforcement Learning: 3D Object Pushing}

\begin{figure}[thpb]
  \centering
  \includegraphics[width=1.0\linewidth]{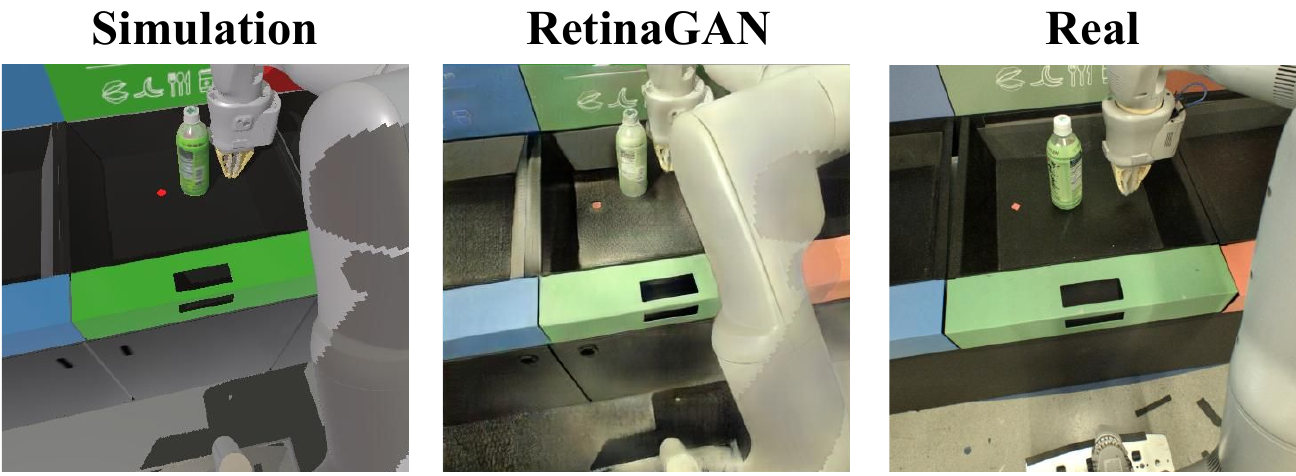}
  \caption{Example unpaired images from the object pushing task, where the robot needs to push an upright object to the goal position, the red dot, without knocking it over.}
  \label{fig:pushing}
\end{figure}

We investigate the transfer capability of RetinaGAN within the same sorting station environment by solving a 3D object pushing task. We test the same RetinaGAN model with this visually similar but distinct robotic pushing task and show that it may be reused without fine-tuning. No additional real data is required for both the pushing task and RetinaGAN.

The pushing task trains purely in simulation, using a scene with a single bottle placed within the center bin of the sorting station and the same Q2-Opt RL framework (Fig. \ref{fig:pushing}). Success is achieved when the object remains upright and is pushed to within 5 centimeters of the goal location indicated by a red marker. We stack the initial image (with the goal marker) and current RGB image as input. For both sim and real world evaluation, the robot needs to push a randomly placed tea bottle to a target location in the bin without knocking it over. Further details are described in \cite{COCOI}, a concurrent submission.
\begin{table}[h]
\caption{Success rate mean and estimated standard deviation (est. std.) of pushing an upright tea bottle to goal position across 10 attempts.}
\label{tab:rl_push}
\begin{center}
\begin{tabular}{lcc}
\toprule
\textbf{Model} & \textbf{Push Success} & \textbf{Est. Std.} \\
\midrule
Sim-Only & 0.0\% & 0.0\% \\
RetinaGAN  & \textbf{90.0\%} & 10.0\% \\
\bottomrule
\end{tabular}
\end{center}
\end{table}

Evaluation results are displayed in Table \ref{tab:rl_push}. %  needed for RetinaGAN.
We train a Q2-Opt policy to perform the pushing task in simulation only and achieve 90\% sim success. When deploying the sim-only RL policy to real, we get 0\% success, revealing a large sim-to-real gap. By applying RetinaGAN to the RL training data, we create a policy achieving 90\% success, demonstrating strong transfer and understanding of the real domain.

\subsection{Imitation Learning: Door Opening}
\label{sec:door_opening}

\begin{figure}[thpb]
  \centering
  \includegraphics[width=1.0\linewidth]{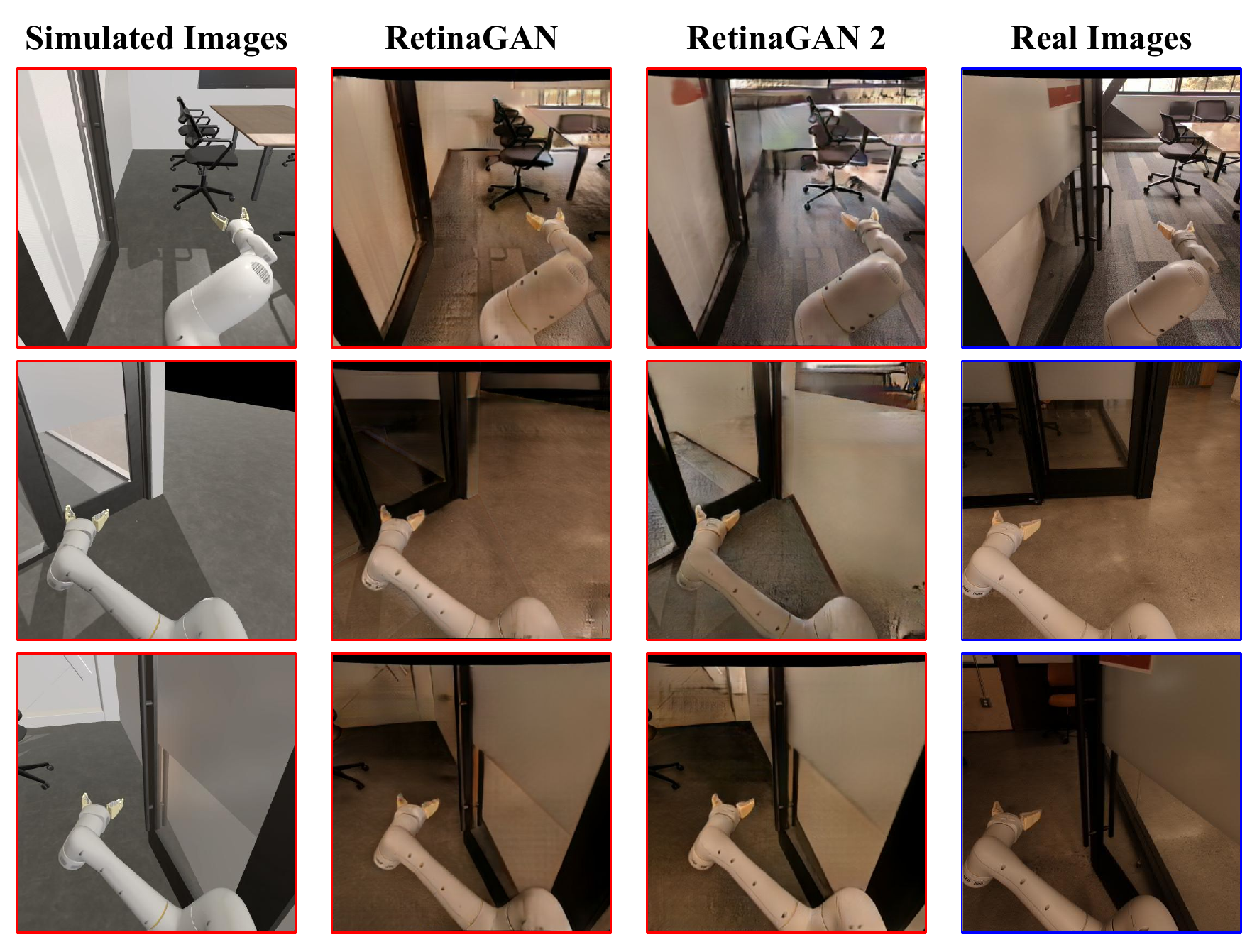}
  \caption{Images sampled from the door opening task in simulation (red border) and real (blue border). Generated images from two separately trained RetinaGAN models highlight prediction diversity in features like lighting or background; this diversity is also present in the real world dataset.}
  \label{fig:door_opening}
\end{figure}

We investigate RetinaGAN with a mis-matched object detector (trained on recycling objects) on an door opening task using a supervised learning form of behavioral cloning and imitation learning (IL). This task is set in a dramatically different visual domain, policy learning framework and algorithm, and neural network architecture. It involves a fixed, extended robot arm with a policy controlling the wheels of the robot base to open the doors of, and enter, conference rooms (Fig. \ref{fig:door_opening}).

The supervised learning policy is represented by a ResNet-FiLM architecture with 18 layers \cite{film}. %(See Appendix for more details).
Both the RetinaGAN model and the supervised learning policy are trained on 1,500 human demonstrations in simulation and 29,000 human demonstrations on real conference doors. We evaluate on three conference rooms seen within the training demonstrations. We train and evaluate on three conference rooms with both left and right-swinging doors, for ten trials each and thirty total trials.

\begin{table}[h]

\caption{Success rate mean and estimated standard deviation (est. std.) of door opening across 30 trials. RetinaGAN+Real result was selected from best of three models used in Multi-RetinaGAN+Real.
}

\label{tab:lfd}
\begin{center}

\begin{tabular}{lcc}

\toprule
\textbf{Model} & \textbf{Seen Doors} & \textbf{Est. Std.} \\
\midrule
Sim-only         & 0.0\% & 0.0\% \\
Real               & 36.6\%  & 8.9\% \\
Sim+Real           & 75.0\%  & 8.0\% \\
RetinaGAN+Real & 76.7\% & 7.9\% \\
Ensemble-RetinaGAN+Real & 93.3\% & 4.6\% \\
Ensemble-RetinaGAN      & \textbf{96.6\%} & 3.4\% \\
\bottomrule
\end{tabular}
\end{center}
\end{table}

With the door opening task, we explore how our domain adapation method performs in an entirely novel domain, training method, and action space, with a relatively low amount of real data. We train the RetinaGAN model using the same object detector trained on recycling objects. This demonstrates the capacity to re-use labeled robot bounding box data across environments, eliminating further human labeling effort. Within door opening images, the perception model produces confident detections only for the the robot arm, but we hypothesize that
structures like door frames could be maintained by consistency in low-probability prediction regimes.

Compared to baselines without consistency loss, RetinaGAN strongly preserves room structures and door locations, while baseline methods lose this consistency (see Appendix).
This semantic inconsistency in GAN baselines presents a safety risk in real world deployment, so we did not attempt evaluations with these models.

We then evaluate IL models trained with different data sources and domain adaptors, and displayed the results in Table \ref{tab:lfd}. An IL model trained on demonstrations in simulation and evaluated in simulation achieves 98\% success. The same model fails in real with no success cases - showing a large sim-to-real gap.

By mixing real world demonstrations in IL model training, we achieve 75\% success on conference room doors seen in training time. We achieve a comparable success rate, 76.7\%, when applying RetinaGAN.

By training on data from three separate RetinaGAN models with different random seeds and consistency loss weights (called Ensemble-RetinaGAN), we are able to achieve 93.3\% success rate. In the low data regime, RetinaGAN can oscillate between various reconstructed semantics and ambiguity in lighting and colors as shown in Fig. \ref{fig:door_opening}. We hypothesize that mixing data from multiple GANs adds diversity and robustness, aiding in generalization. Finally, we attempt Ensemble-RetinaGAN without any real data for training the IL model. We achieve 96.6\%, within margin of error of the Ensemble-RetinaGAN+Real result.

\section{Conclusions}

RetinaGAN is an object-aware sim-to-real adaptation technique which transfers robustly across environments and tasks, even with limited real data. We evaluate on three tasks and show 80\% success on instance grasping, a 12 percentage-point improvement upon baselines. Further extensions may look into pixel-level perception consistency or other modalities like depth. Another direction of work in task and domain-agnostic transfer could extend RetinaGAN to perform well in a visual environment unseen at training time.

\section*{Appendix}

\subsection{Alternative Perception Losses}
We note that it is also possible to train separate perception networks for each domain. However, this adds complexity and requires that the object sets between synthetic and real data be close to bijective, because both models would have to produce consistent predictions on perfectly paired images.

Providing perception consistency with off-the-shelf, pre-trained models is a promising future direction that eliminates the costs of perception model creation. Future work may investigate whether such models can be successfully leveraged to train RetinaGAN. As they are likely trained solely on real data, the relatively unbalanced predictions between the sim and real domains may destablize training.

While segmentation models like Mask-RCNN \cite{he2018mask} and ShapeMask \cite{kuo2019shapemask} provide dense, pixel-level object supervision, it is practically easier and more efficient to label object detection data. % Existing detection labeled data and models are more common, and data collection and labeling efforts can be amortized across various use cases.
However, it may provide a stronger supervision signal, and semantic segmentation models may provide stronger consistency for background objects and structures.

\subsection{Door Opening Figure}
\begin{figure}[thpb]
  \centering
  \includegraphics[scale=0.6]{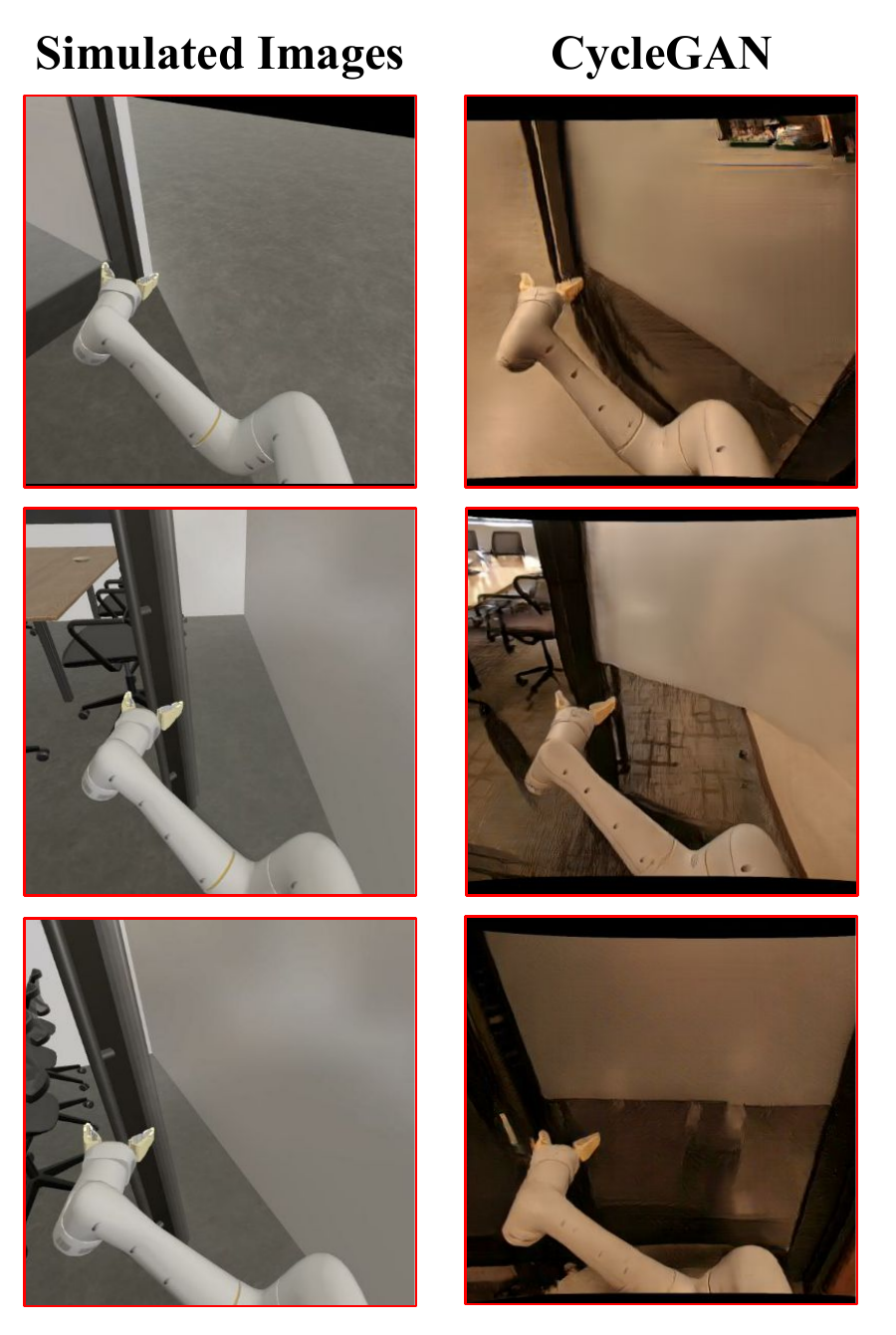}
  \caption{CycleGAN can distort semantic structure when trained on door opening images, in the low data regime. Images on the right are transfered results of the simulated image on the left.}
  \label{fig:gan4}
\end{figure}

See Fig. \ref{fig:gan4} for example of semantic structure distortions when training the door opening task with CycleGAN.

\subsection{Perception Model Training}
Hyperparameters used in object detection model training are listed in Table \ref{tab:prcp_hparam}. We use default augmentation parameters from \cite{retinanet}, including a scale range of 0.8 to 1.2. Among the 59 classes, the following are frequently used: robot, bottle, bowl, can, cup, bag/wrapper, bowl, and plate. Other classes appear sparesely or not at all.

\begin{table}[H]
\centering
\caption{Hyperparameters used for EfficientDet Training.}
\label{tab:prcp_hparam}
\begin{center}
\begin{tabular}{l c}
\toprule
\textbf{Hyperparameter} & \textbf{Value} \\
\midrule
Training Hardware & 4 x Google TPUv3 Pods \\
Network Architecture & EfficientNet-D1 \cite{efficientdet} \\
Precision & \texttt{bfloat16} \\
Input Resolution & 512x640 pixels \\
Preprocessing   & Crop, scale, Horizontal flipping \\
              & Pad to 640x640 \\
Training Step Count & 90,000 \\
Optimizer & \texttt{tf.train.MomentumOptimizier} \\
Learning Rate & 0.08, stepped two times with 10\% decay  \\
Momentum & 0.08  \\
Batch Size & 256 \\
Weight Decay & 1e-5 \\
Classes & 59 \\
\bottomrule
\end{tabular}
\end{center}
\end{table}

\subsection{RetinaGAN Model Training}
\label{appendix:retinagan}

We train RetinaGAN following the hyper-parameters described in Appendix A of \cite{rao2020rlcyclegan}. We did not tune any CycleGAN hyper-parameters, and we primarily searched between $0.1$ and $1$ for $\mathcal{L}_\text{prcp}$. We did not run any hyper-parameter search on relative weights between $\mathcal{L}_\text{prcp}$ terms. We generate simulation images with the following object set (and counts): paper bags (1), bottles (9), bowls (1), napkins (1), cans (12), cups (6), containers (2), plates (1), and wrappers (10). Each training batch includes 256 simulation and 256 real images. Photometric distortions are defined in the Tensor2Robot framework\footnote{\url{https://github.com/google-research/tensor2robot/blob/master/preprocessors/image_transformations.py}}.

\begin{table}[H]
% \centering
\caption{Hyperparameters used for GAN Training.}
\label{rl_module_table}
% \begin{center}
\begin{tabular}{l c}
\toprule
\textbf{Hyperparameter} & \textbf{Value} \\
\midrule
Training Hardware & 4 x Google TPUv3 Pods \\
Network Architecture & U-Net \cite{ronneberger2015unet}, Fig. 5 in \cite{rao2020rlcyclegan} \\
Precision & \texttt{bfloat16} \\
Input Resolution & 512x640 pixels \\
Preprocessing & Crop to 472x472 pixels \\
              & Apply photometric distortions \\
Training Step Count & 50,000-100,000 \\
Optimizer & \texttt{tf.train.AdamOptimizer} \\
          & $\beta_1=0.1, \beta_2=0.999$ \\
Learning Rate & 0.0001 \\
Batch Size & 512 \\
Weight Decay & 7e-5 \\
Additional Normalization & Spectral Normalization \cite{zhang2019selfattention} \\
$\mathcal{L}_\text{GAN}$ & 1 weight ($\lambda_\text{GAN}$), updates $G, F, D_x, D_y$ \\
$\mathcal{L}_\text{cycle}$ & 10 weight ($\lambda_\text{cycle}$), updates $G, F$ \\
$\mathcal{L}_\text{prcp}$ & 0.1 weight ($\lambda_\text{prcp}$), updates $G, F$ \\
\bottomrule
\end{tabular}
% \end{center}
\end{table}

\subsection{Q2-Opt RL Model Training}
We use the Q2R-Opt \cite{bodnar2019quantile} model and training pipeline for both the grasping and pushing tasks. We use the same hyper-parameters as in this prior work, without any tuning. We train on the same simulated object set as in the RetinaGAN setup.

When using the full real dataset, we sample each minibatch from simulation episodes with a 50\% weight and real episodes with a 50\% weight. With the restricted 10K episode dataset, we sample from simulation with 20\% weight and real with 80\% weight, as to not overfit on the smaller real dataset.
We did not tune these ratios, as in prior experiments, we found that careful tuning was not required.

\subsection{ResNet-FiLM IL Model Training}
We train IL with the ResNet-FiLM \cite{film} model with a ResNet-18 architecture defined in the Tensor2Robot framework\footnote{\url{https://github.com/google-research/tensor2robot/blob/master/layers/film_resnet_model.py}}. For training RetinaGAN and Multi-RetinaGAN, we mix real demonstrations, simulated demonstrations, and RetinaGAN-adapted simulated demonstrations. We use a lower 20\% weight for real data (because of the small dataset size) and evenly weight simulated and adapted demonstrations. The action space is the 2D movement of the robot base. Additional details will be provided in an as-yet unreleased paper; this work focuses on the benefits of CycleGAN-adapted data independently of whether policies are trained with IL or RL. We used the same hyper-parameters for all experiments.

\subsection{Evaluation}
\begin{figure}[thpb]
  \centering
  \includegraphics[scale=0.6]{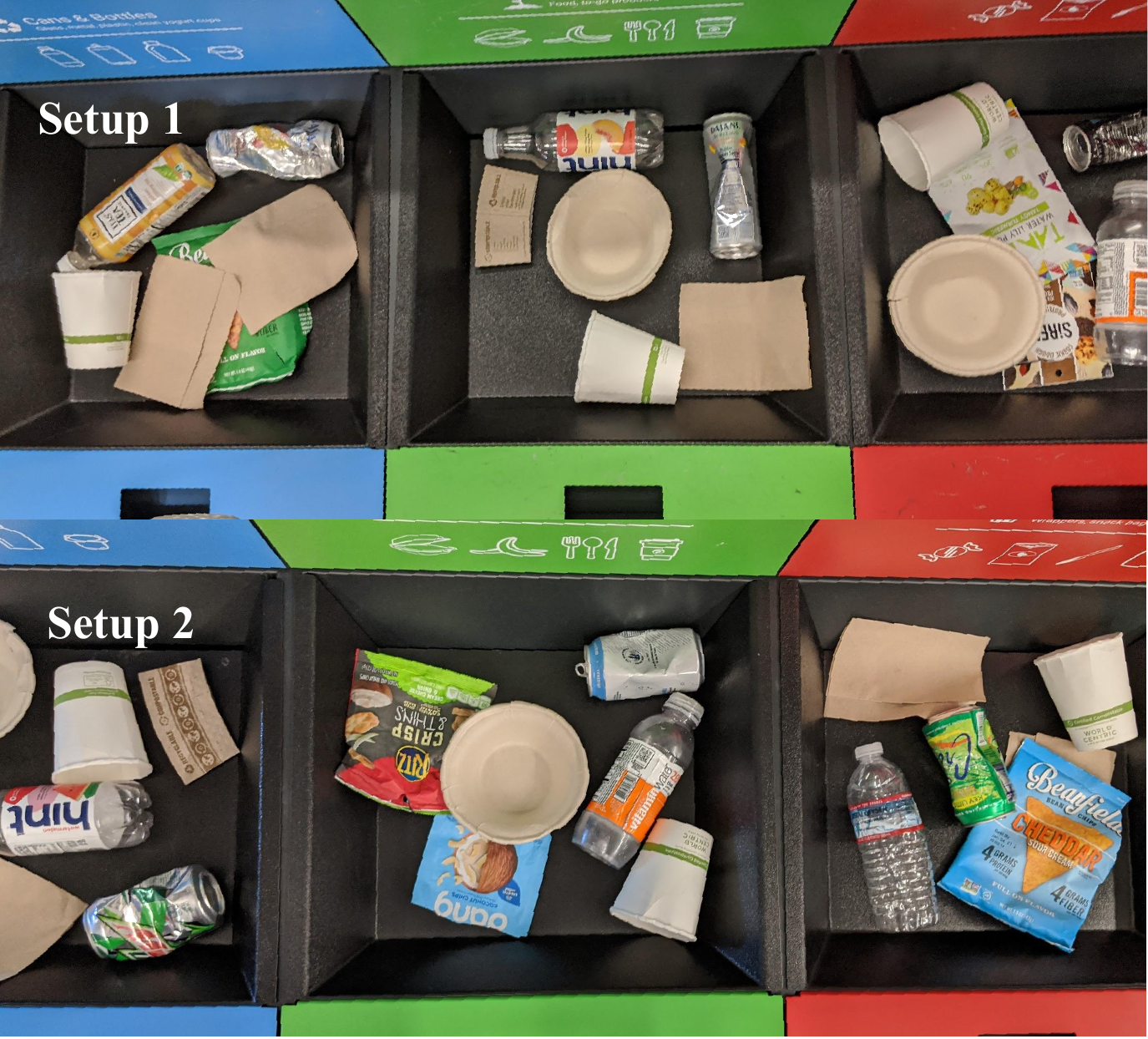}
  \caption{The two evaluation station setups displaying the object classes present in each bin.}
  \label{fig:eval}
\end{figure}

For grasping, we evaluate with the station setup in Fig. \ref{fig:eval}. Each setup is replicated three times (with potentially different object brands/instances, but the same classes), and one robot positioned in front of each bin. We target the robot to only grasp the cup, can, and bottle, for a total of eighteen grasps. This is repeated five times for ninety total grasps. 

For pushing, we evaluate with a single Ito En Green Tea Bottle filled 25\% full of water.

For door opening, we evaluate on three real world conference room doors. Two doors swing rightwards and one door swings leftwards. The episode is judged as successful if the robot autonomously pushes the door open and the robot base enters the room.

% \newpage

%%%%%%%%%%%%%%%%%%%%%%%%%%%%%%%%%%%%%%%%%%%%%%%%%%%%%%%%%%%%%%%%%%%%%%%%%%%%%%%%

\section*{Acknowledgements}

We thank Noah Brown, Christopher Paguyo, Armando Fuentes, and Sphurti More for overseeing robot operations, and Daniel Kappler, Paul Wohlhart, and Alexander Herzog for helpful discussions. We thank Chris Harris and Alex Irpan for comments on the manuscript.

%%%%%%%%%%%%%%%%%%%%%%%%%%%%%%%%%%%%%%%%%%%%%%%%%%%%%%%%%%%%%%%%%%%%%%%%%%%%%%%%
\newpage
% \addtolength{\textheight}{-17cm}
\bibliographystyle{bib/IEEEtran} % use IEEEtran.bst style
\bibliography{bib/IEEEabrv,root}

\end{document}